\documentclass[preprint,review]{elsarticle}
\usepackage{graphicx}
\usepackage{subfigure}
\usepackage{amsfonts}
\usepackage{amsmath}
\usepackage{amssymb}
\usepackage{rotating}
\usepackage{xr} 
\usepackage[ruled]{algorithm2e}
\usepackage{algorithmic}
\usepackage{url}

\begin{document}

\begin{frontmatter}

\title{3D Face Recognition with Sparse Spherical Representations\tnoteref{thanks}}
\tnotetext[thanks]{This work has been partly supported by the
Swiss National Science Foundation, under grants NCCR IM2 and
200020-120063.}

\author[Barcelona_address]{R. Sala Llonch}
\author[EPFL_address]{E. Kokiopoulou}
\author[EPFL_address]{I. Tosic}
\author[EPFL_address]{P. Frossard}

\address[Barcelona_address]{Hospital Clinic - Universitat de Barcelona,
08028 Barcelona, Spain.}

\address[EPFL_address]{Signal Processing Laboratory (LTS4),
Ecole Polytechnique F\'ed\'erale de Lausanne (EPFL), Lausanne
1015, Switzerland.}

\begin{abstract}
This paper addresses the problem of 3D face recognition using
simultaneous sparse approximations on the sphere. The 3D face
point clouds are first aligned with a novel and fully automated
registration process. They are then represented as signals on the
2D sphere in order to preserve depth and geometry information.
Next, we implement a dimensionality reduction process with
simultaneous sparse approximations and subspace projection. It
permits to represent each 3D face by only a few spherical
functions that are able to capture the salient facial
characteristics, and hence to preserve the discriminant facial
information. We eventually perform recognition by effective
matching in the reduced space, where Linear Discriminant Analysis
can be further activated for improved recognition performance. The
3D face recognition algorithm is evaluated on the FRGC v.1.0 data
set, where it is shown to outperform classical state-of-the-art
solutions that work with depth images.
\end{abstract}


\end{frontmatter}

\section{Introduction}\label{sec:intro}

Automatic recognition of human faces is an actively researched
area, which finds numerous applications such as surveillance,
automated screening, authentication or human-computer interaction.
The face is an easily collectible, universal and non-intrusive
biometric \cite{Jain}, which makes it ideal for applications where
other biometrics such as fingerprints or iris scanning are not
possible.

There has been a considerable progress in the area of
two-dimensional face recognition where intensity/color images of
human faces are employed. However, these systems are sensitive to
illumination, pose variations, occlusions, facial expressions and
make-up. On the other hand, recognition systems based on 3D face
information have the potential for greater recognition accuracy
and are capable of overcoming part of the limitations of 2D face
recognition systems \cite{Bowyer,Akarun}. The 3D shape of a face,
usually given as a 3D point cloud, depends on its anatomical
structure and it is independent of its pose, which can be further
corrected by rigid rotations in the 3D space \cite{Besl}.

\begin{figure}[t]
\begin{center}
  \includegraphics[width=0.45\textwidth]{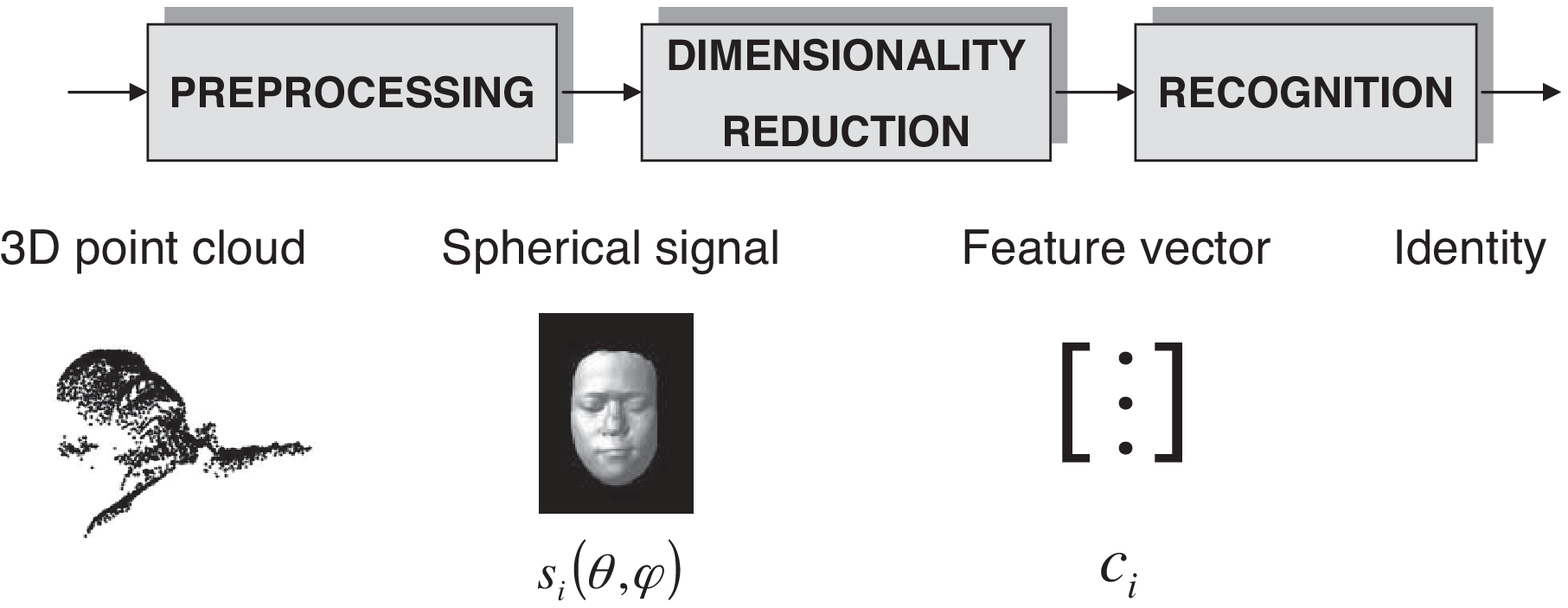}
  \caption{Block diagram of the 3D face recognition system.}
  \label{fig:algorithm}
\end{center}
\end{figure}

We consider in this paper the problem of 3D face recognition and
we design a fully automatic algorithm based on simultaneous sparse
expansions on the sphere. We first propose a preprocessing step
that automatically registers the 3D point clouds prior to
dimensionality reduction. It selects the facial region and
registers all the faces by an accurate automatic two-step
algorithm based on an Average Face Model (AFM) and on the
Iterative Closest Point (ICP) algorithm \cite{Besl}. Contrarily to
most of the existing algorithms, the proposed registration process
does not require any manual intervention. Registered point clouds
are then mapped on the 2D sphere where the spherical face
functions are created by nearest neighbor interpolation. The
spherical representation enables the use of spherical signal
processing techniques, which consider the face signals as
combinations of basis functions with diverse shape, position and
orientation on the sphere.

The spherical face signals then undergo a dimensionality reduction
step that represents each face with a reduced set of discriminant
features. We build a dictionary of functions on the sphere and we
select the discriminant basis functions by simultaneous sparse
approximations. The face signals are finally projected onto the
resulting reduced subspace, in order to generate feature vectors.
We finally implement a recognition step where Linear Discriminant
Analysis (LDA) is performed on the subspace representation of the
faces. The recognition system is illustrated on Fig.
\ref{fig:algorithm}, where $s_i (\theta, \phi)$ denotes the
spherical signal $s_i$ as a function of position $(\theta, \phi)$
on the 2D sphere, and $c_i$ is a feature vector.

The performance of the 3D face recognition system is evaluated on
the FRGC v.1.0 data set. The proposed algorithm outperforms
state-of-the-art solutions based on Principal Component Analysis
(PCA, \cite{PCA}) or Linear Discriminant Analysis (LDA) on depth
images. Our fully automatic system provides effective
classification performance that shows that 3D face recognition
with spherical representations certainly represents a promising
solution for person identification.

The paper is organized as follows. We provide an overview of the
related work in 3D face recognition in Section II. Section III
describes the automatic face registration process that permits to
align the 3D points clouds before analysis. The dimensionality
reduction step with simultaneous sparse approximations on the
sphere is presented in Section IV and experimental results are
finally provided in Section V.

\section{Related work}\label{sec:related}

3D face recognition has attracted a lot of research efforts in the
past few decades due to the advent of new sensing technologies and
the high potential of 3D methods for building robust systems with
invariance to head pose and illumination variations. We review in
this section the most relevant work in 3D face recognition, which
can be categorized in methods using point cloud representations,
depth images, facial surface features or spherical representations
respectively. Surveys of the state-of-the-art in 3D face
recognition are further provided in \cite{Bowyer,Akarun}.

The recognition methods that work directly on 3D point clouds
consider the data in their original representation based on
spatial and depth information. A priori registration of the point
clouds is commonly performed by ICP algorithms
\cite{Besl,Gokberk}. The classification is generally based on the
Hausdorff distance that permits to measure the similarity between
different point clouds \cite{Achermann}. Alternatively,
recognition could be performed with ``3D eigenfaces'' that are
constructed directly from the 3D point clouds \cite{Xu}. The main
drawback of the recognition methods based on 3D point clouds
however resides in their high computational complexity that is
driven by the large size of the data.

Many recognition systems use depth or range images that permit to
formulate the 3D face recognition as a problem of dimensionality
reduction for planar images, where each pixel value represents the
distance from the sensor to the facial surface. Principal
Component Analysis (PCA) and ``Eigenfaces'' can be used for
dimensionality reduction \cite{Turk}, where the basis vectors are
however typically holistic and of global support. PCA can be
combined with Linear Discriminant Analysis (LDA) to form
``Fisherfaces'' with enhanced class separability properties
\cite{Belhumeur}. Alternatively, dimensionality reduction can be
performed via variants of non-negative matrix factorization (NMF)
algorithms \cite{Lee,Paatero,HoyerNMF} that produce part-based
decompositions of  the depth images. Part-based decompositions
based on non-negative sparse coding \cite{HoyerNNSC} have recently
been shown to provide improved recognition performance than NMF
methods in face recognition \cite{Shastri}. Recent methods have
proposed to concentrate dimensionality reduction around facial
landmarks like the nose tip \cite{Jahanbin} or in multiple
carefully chosen regions \cite{Wong} or to compute geodesic
distances among the selected fiducial points \cite{Gupta}. They
however require a selection of the fiducial points or areas of
interest that is often performed manually and prevents the
implementation of fully automatic systems.

Facial surface features have also been proposed for 3D face
recognition. The idea of recognizing 3D faces using curvature
descriptors has been originally introduced in \cite{Gordon}, where
features are chosen to represent both curvature and metric size
properties of faces. More recently, level sets of the depth
function on range image have been used to define sets of facial
curves \cite{Samir}. They are further embedded in an appropriately
defined shape manifold and compared based on geodesic distances.
Facial curve representations provide global information about the
whole facial surface, which unfortunately does not permit to take
advantage of discriminative local features.

Finally, spherical representations have been used recently for
modelling illumination variations \cite{Wang,Ramamoorthi} or both
illumination and pose variations in face images \cite{Yue}.
Spherical representations permit to efficiently represent facial
surfaces and overcome the limitations of other methods towards
occlusions and partial views \cite{Hebert}. To the best of our
knowledge, the representation of 3D face point clouds as spherical
signals for face recognition has however not been investigated
yet. We therefore propose to take benefit of the robustness of
spherical representations and of spherical signal processing tools
to build an effective and automatic 3D face recognition system. We
perform dimensionality reduction directly on the sphere, so that
the geometry of 3D faces is preserved. The reduced feature space
is extracted by sparse approximations with a dictionary of
localized geometric features on the sphere that effectively
capture spatially localized and salient 3D face features that are
advantageous in the recognition process.

\section{Automatic preprocessing of 3D face data}
\label{sec:prepro}

\subsection{Automatic face extraction}\label{sec:faceextract}

\begin{figure}[t]
\centering \subfigure[Binary matrix $A$]{
    \frame{\includegraphics[width=.21\textwidth]{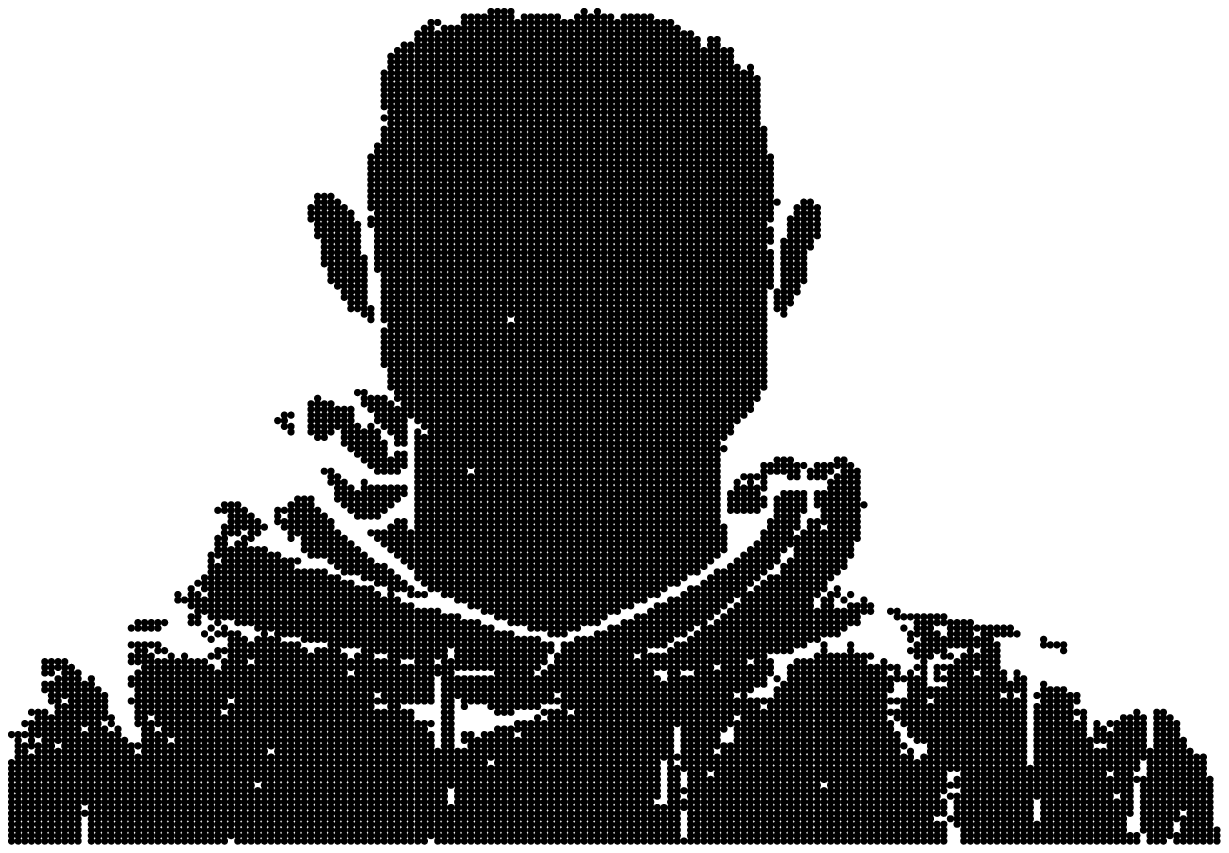}\label{fig:prepro_a}}}
\subfigure[After lateral thresholding]{
    \frame{\includegraphics[width=.21\textwidth]{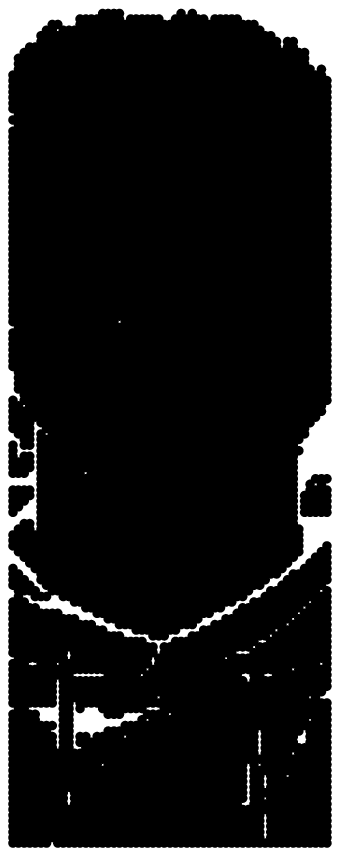}\label{fig:prepro_b}}}
    \\
\subfigure[Profile view]{
    \frame{\includegraphics[width=.21\textwidth]{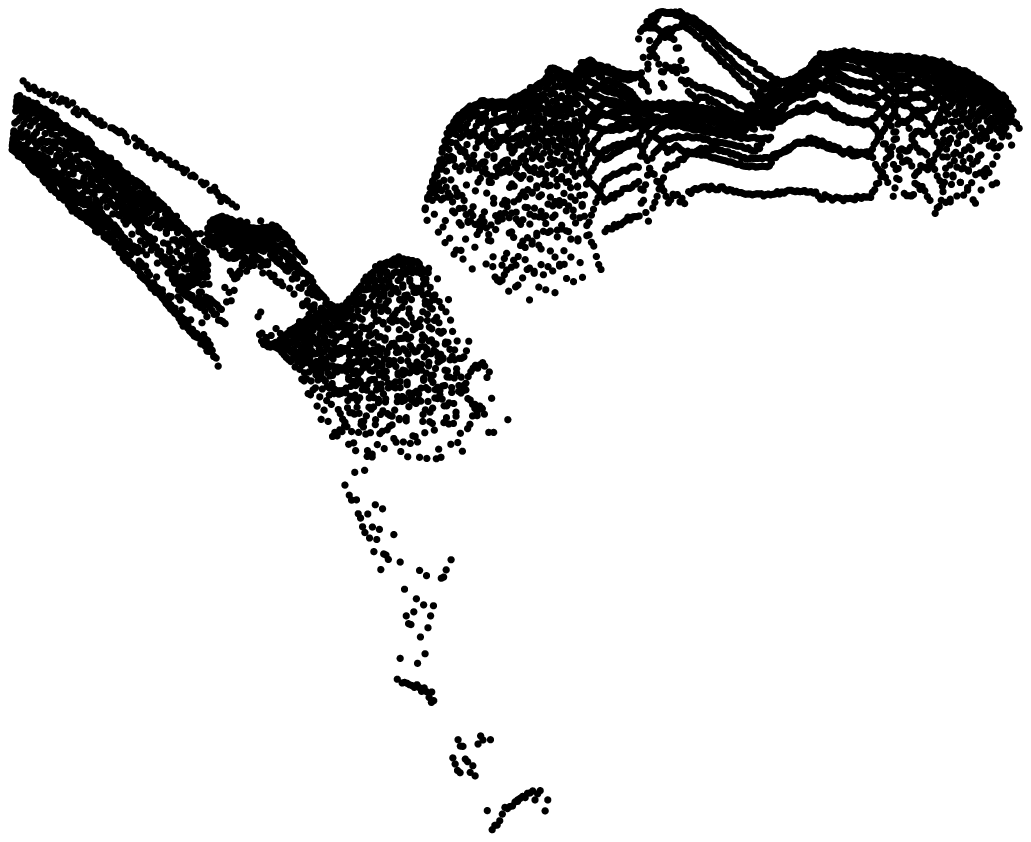} \label{fig:perfil_a}}}
\subfigure[After depth thresholding (profile view)]{
    \frame{\includegraphics[width=.21\textwidth]{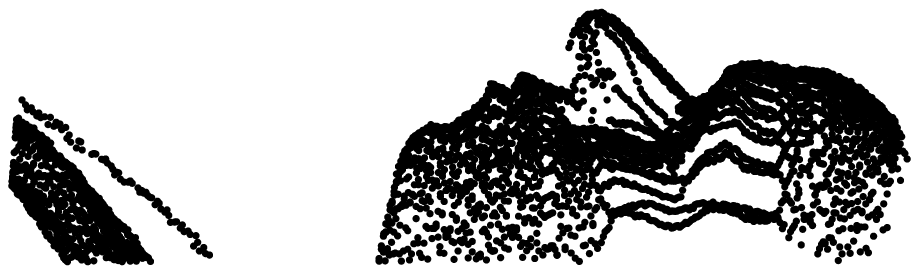}\label{fig:perfil_b}}}
    \\
\subfigure[After depth thresholding]{
    \frame{\includegraphics[width=.21\textwidth]{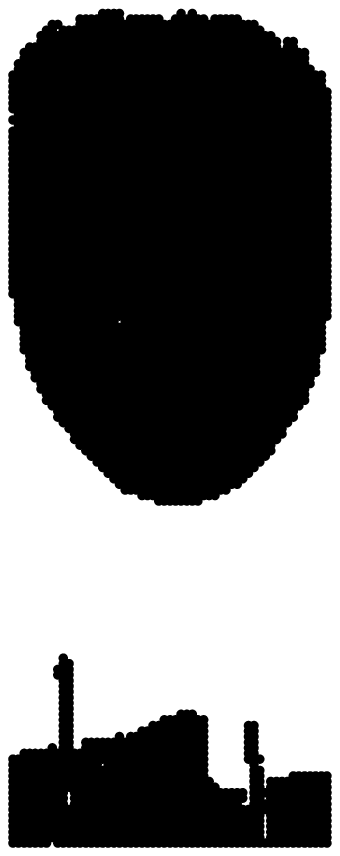}\label{fig:prepro_c}}}
\subfigure[After morphological processing]{
    \frame{\includegraphics[width=.21\textwidth]{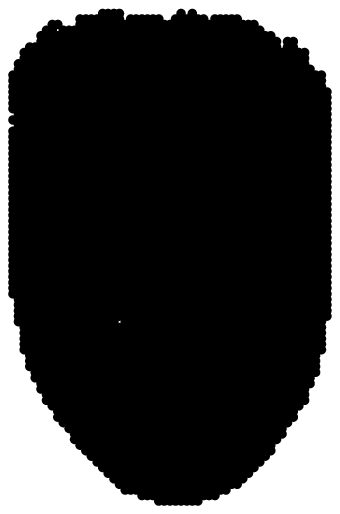}\label{fig:prepro_d}}}
\caption{Main steps in facial region extraction}
\end{figure}

We propose in this section a fully automatic preprocessing method
for preparing and aligning 3D face point clouds before feature
extraction and recognition. Unlike most of the algorithms in the
literature, the preprocessing step does not require any manual
intervention, which is an enormous advantage for the design of
fully automated face recognition systems. The preprocessing scheme
is based on two main tasks, respectively the extraction of the
facial region, and the registration of the 3D face. We present
these tasks in more details in the rest of the section.

The main purpose of the face extraction step is to remove
irrelevant information from the 3D point clouds, such as data that
correspond to shoulder, or hair for example. The output of a
facial scan typically forms a 3D point cloud $\{X,Y,Z\}$, where
$X$ and $Y$ form a uniform Euclidean grid and $Z$ provides the
corresponding depth values. The point cloud is also accompanied by
a binary matrix $A$ of valid points, which has the same resolution
as the grid implied by $X \times Y$. The nonzero pattern of such a
sample binary matrix is shown in Fig. \ref{fig:prepro_a}. There is
however no guarantee that the points exclusively correspond to
face depth information, and face extraction is therefore necessary
to ensure that the feature extraction concentrates on capturing
discriminative facial information.

The first step in face extraction consists in removing data points
on the subject's shoulders. We estimate a vertical projection
curve from the point cloud by computing the column sum of the
matrix $A$. Then, we define two lateral thresholds on the left and
right inflexion points of the projection curve, and we remove all
data points beyond these thresholds, as illustrated in Fig.
\ref{fig:prepro_b}. We further remove the data points
corresponding to the subject's chest by thresholding of the
histogram of depth values. It removes the data points with large
depth values that are typically situated behind the data
corresponding to frontal face information, as shown in Figs
\ref{fig:perfil_a} and \ref{fig:perfil_b}. We finally have to
remove outlier points that remain in regions disconnected from the
main facial area, as shown in Fig. \ref{fig:prepro_c}. We
therefore perform morphological image processing on the
corresponding binary matrix $A$, where we keep only the largest
region that typically correspond to the facial region, as
presented in \ref{fig:prepro_d}.

\subsection{Automatic face registration}\label{sec:facereg}

After extracting the main facial region from the 3D scans, the
face signals have to be registered in order to ensure that all
have the same pose before the recognition step. The registration
typically applies rigid transformations on the 3D faces in order
to align them. We propose a two-step approach for automatic
registration, where an Average Face Model (AFM) is computed and
then used for accurate registration.

\begin{figure}[t]
\centering \subfigure[Depth map]{
    \includegraphics[width=.22\textwidth]{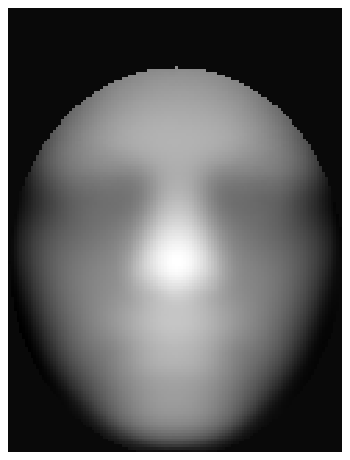} \label{fig:afm_a}}
\subfigure[Point cloud]{
    \includegraphics[width=.22\textwidth]{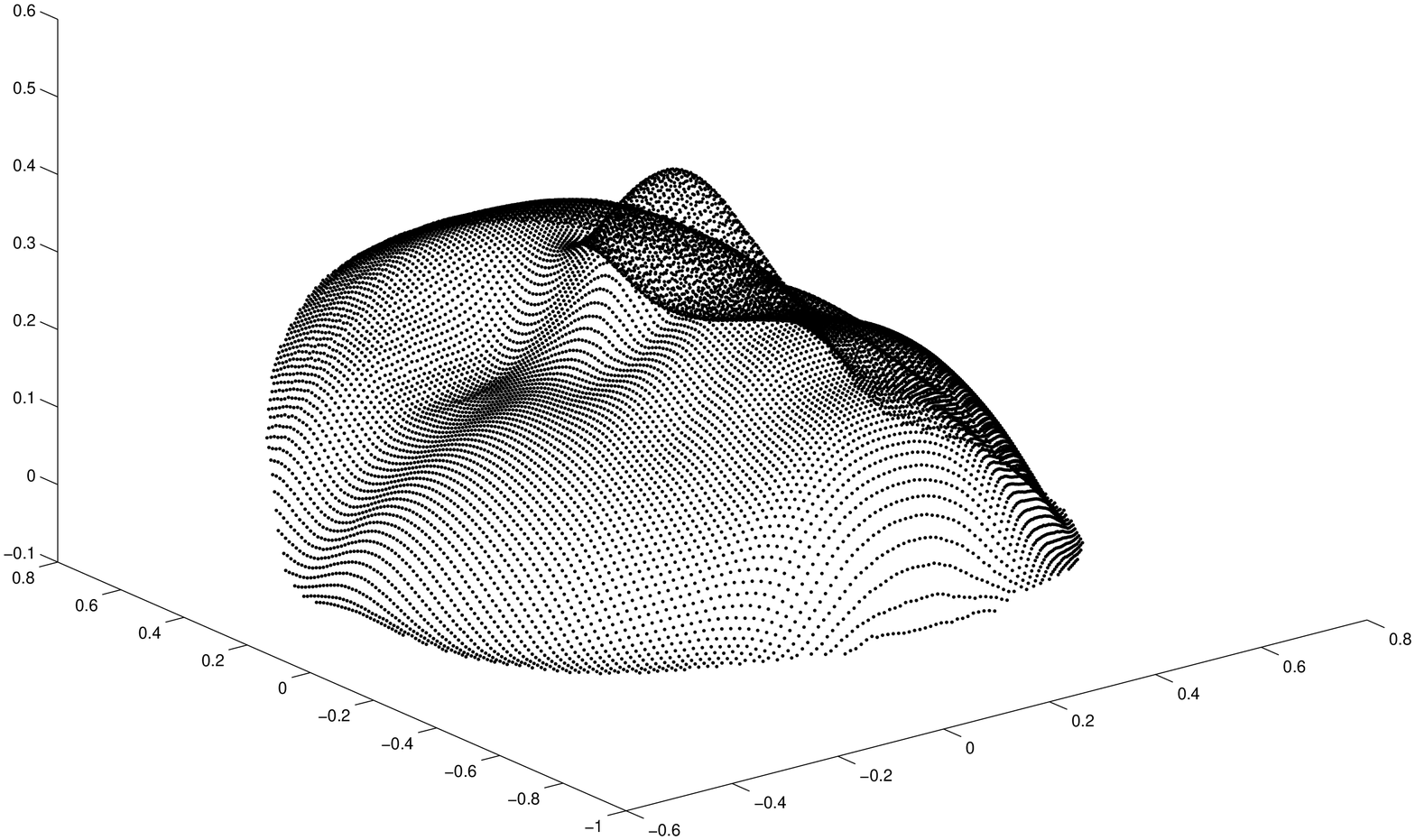}\label{fig:afm_b}}
\caption{Average Face Model given as a depth map or a 3D point
cloud.} \label{fig:afm1}
\end{figure}

First, we randomly pick a training face, and we align all the
faces approximately to the sample face using the Iterative Closest
Point (ICP) algorithm \cite{Besl}. Given a model and a query point
cloud, ICP computes a rigid transformation, consisting of
rotations and translations, by minimizing the sum of square errors
between the closest model points and query points. After coarse
registration with ICP, the face signals are re-sampled on a
uniform 2D grid using nearest neighbor interpolation. It permits
to construct an AFM, by computing at each grid point the average
depth value among all training faces (see Figure \ref{fig:afm1}) .
The AFM is subsequently used as reference in order to define an
ellipse that contains the main facial region. Since, the faces are
already registered, this ellipse can be used to crop closely all
faces in the training set. The ellipse cropping step removes all
the irrelevant information that may be left over from the previous
preprocessing steps, as shown in Figure \ref{fig:ellipse
cropping}.

A fine alignment of the faces can now be performed on the signals
that have been cleaned from outliers. The accurate alignment is
finally obtained by running ICP one more time. The AFM is now used
as a reference face model, and all faces signals are registered
with respect to the AFM.

\begin{figure}[t]
\centering \subfigure[Before (depth map)]{
    \includegraphics[width=.22\textwidth]{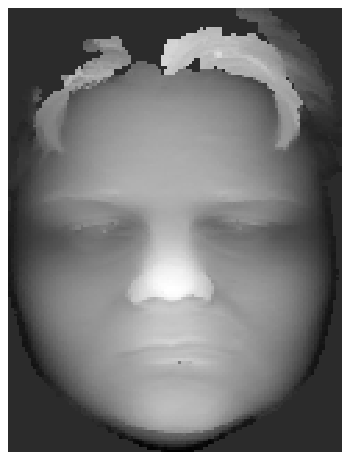} \label{fig:afm_c}}
\subfigure[Before (point cloud)]{
    \includegraphics[width=.22\textwidth]{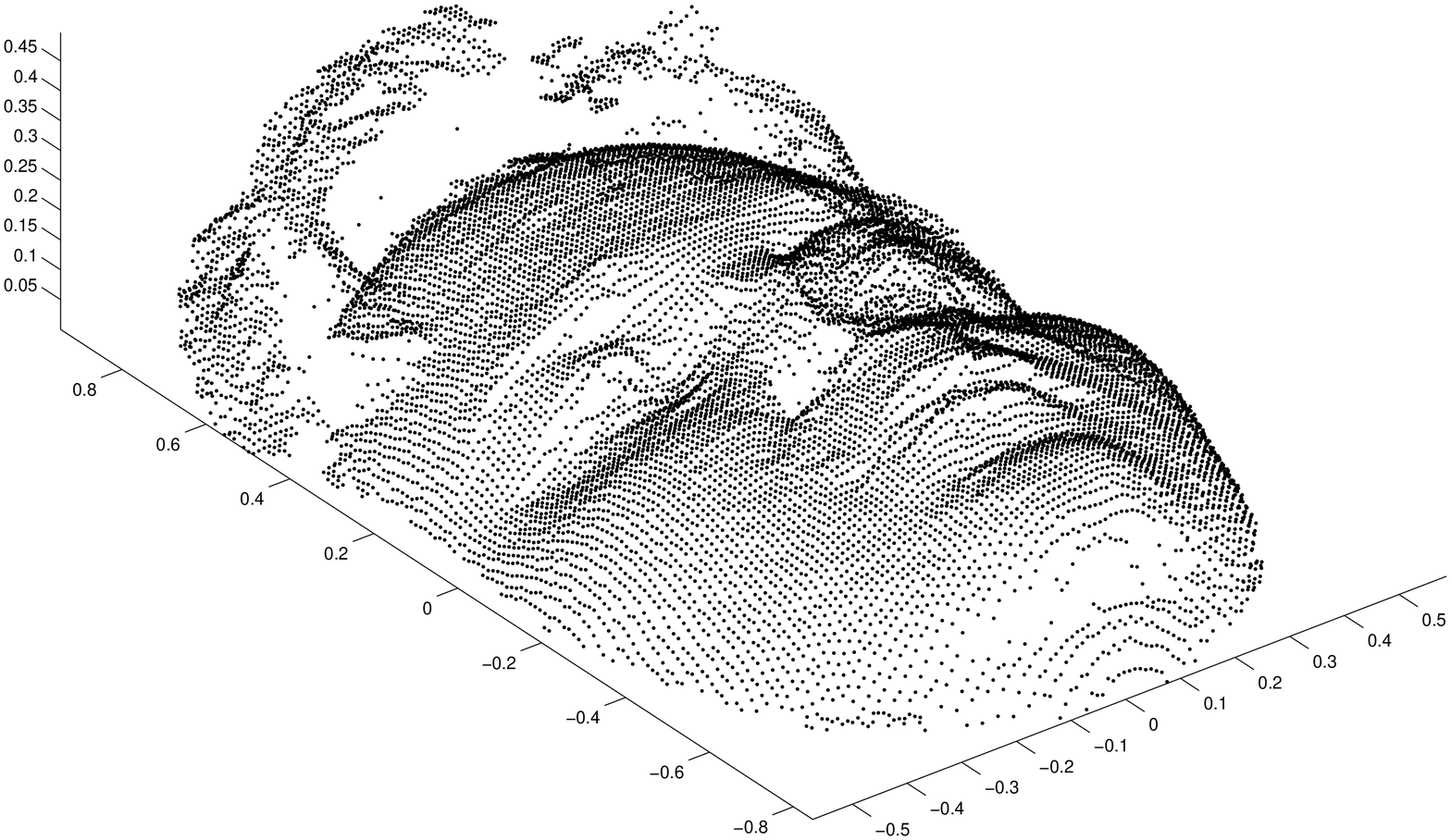}\label{fig:afm_d}}\\
\subfigure[After (depth map)]{
    \includegraphics[width=.22\textwidth]{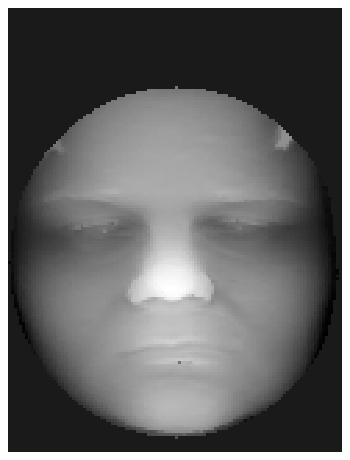} \label{fig:afm_e}}
\subfigure[After (point cloud)]{
    \includegraphics[width=.22\textwidth]{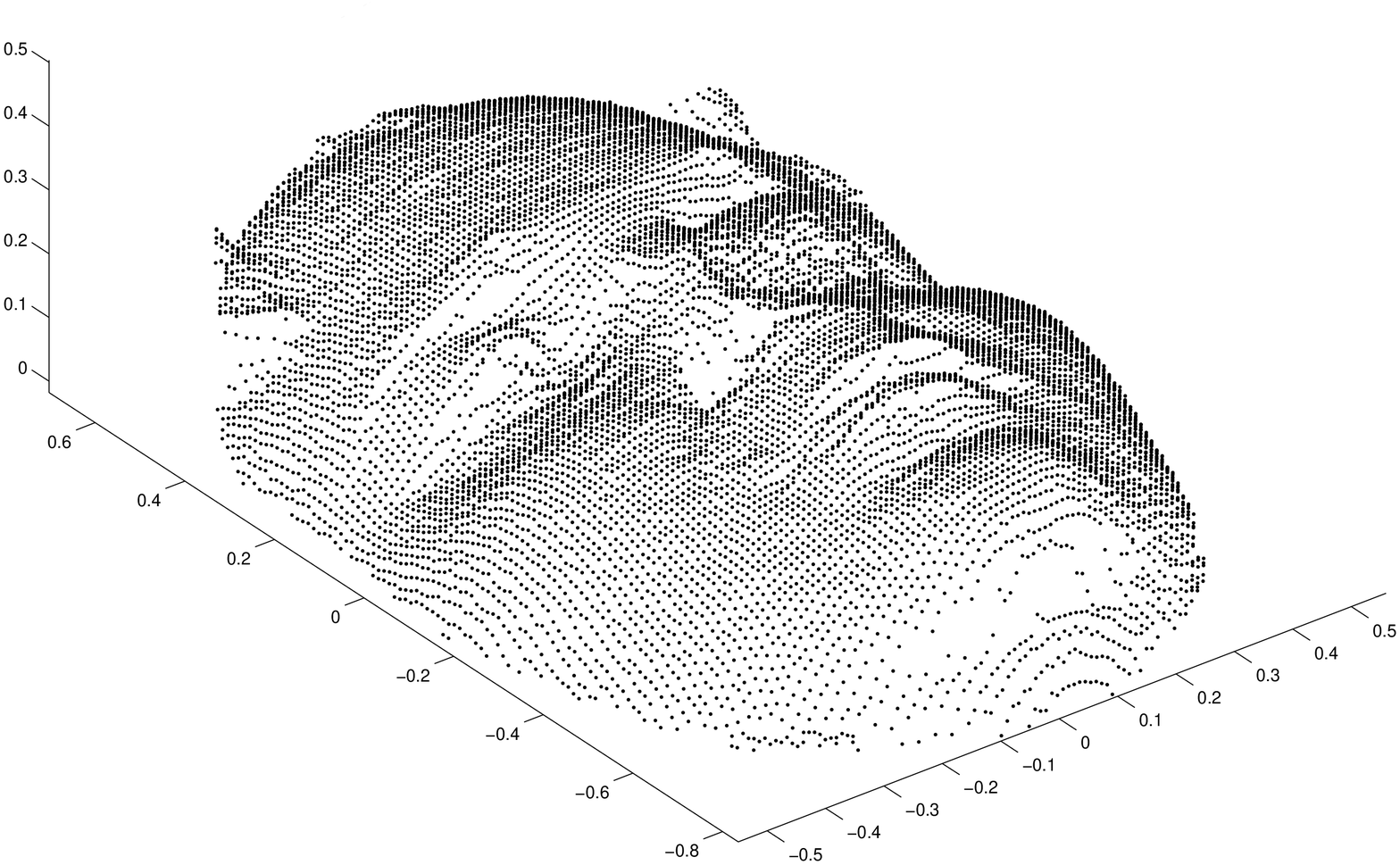}\label{fig:afm_f}}
\caption{Illustration of ellipse cropping on depth maps and equivalent 3D point clouds.} \label{fig:ellipse cropping}
\end{figure}

\section{Recognition with sparse spherical representations}

\subsection{Simultaneous sparse approximations}

Efficient face recognition algorithms usually include a
dimensionality reduction step, where high dimensional data are
represented in a reduced subspace. We propose to use sparse signal
representation methods for dimensionality reduction. Such methods
have demonstrated good performance in 2D face
recognition~\cite{Efi}. They present the advantage of capturing
the main signal characteristics in a very small set of meaningful
features, which are moreover defined a priori in a dictionary of
functions. This presents an interesting advantage compared to
classical methods such as PCA, whose feature vectors are
data-dependent. In addition, a proper choice of the dictionary
permits to build features that capture the geometrical information
in the face signal. We give below a brief overview of sparse
approximations, and we show later how we use them for
dimensionality reduction on the sphere.

Let denote by $s_i, i=1,...,N$, a set of functions in the Hilbert
space $\mathcal{H}$. Let further denote by
$\mathcal{D}=\{g_\gamma,\gamma \in \Gamma\}$ an overcomplete
dictionary of unit $L_2$ norm functions indexed by $\gamma$, which
spans the space $\mathcal{H}$. A function $s_i$ has a sparse
representation in $\mathcal{D}$ if it can be represented in terms
of a linear superposition of small set of basis functions
$\{g_\gamma\} \in \mathcal{D}$ . In other words, it can be
expressed as $s_i = {\Phi_I}_i c_i$, where ${\Phi_I}_i$ denotes a
matrix whose columns are atoms in ${\mathcal{D}_I}_i \subset
\mathcal{D}$ that forms the sparse support of the signal $s_i$.
The vector $c_i$ represents the coefficients of the linear
approximation of $s_i$ with atoms in ${\mathcal{D}_I}_i $.

Finding the sparsest representation of a signal in a redundant
dictionary $\mathcal{D}$ is in general an NP-hard problem. Greedy
algorithms like Matching Pursuit \cite{Mallat} have however shown
to provide suboptimal yet efficient solutions with a limited
computational complexity. It selects iteratively the functions
from the dictionary that best matches the signals $s_i$. We have
however to ensure that the atoms that form the support of the
different signals $s_i$'s are identical, in order to permit to
classify them in the feature space. Dimensionality reduction can
thus be performed by simultaneous decomposition of all the signals
$s_i, i=1,...,N$. Finding the sparse support $\mathcal{D}_I$ that
is common to all the signals $\{s_i\}$ can be achieved by the
Simultaneous MP (SMP)~\cite{Tropp} algorithm, which only induces a
small increase of complexity compared to MP on a single signal
\cite{Efi}. In short, SMP greedily selects $\mathcal{D}_I$ such
that all the $N$ functions $s_i$ are simultaneously approximated
in the same basis. It results in the extraction of $K$ atoms such
that all signals are simultaneously represented by linear
combinations of them. Each signal can be re-written as $s_i =
\Phi_I c_i$, where $\Phi_I$ denotes the matrix whose columns are
the atoms in the common sparse support $\mathcal{D}_I \subset
\mathcal{D}$. Finally, a few iterations are typically sufficient
to capture most of the energy of the face signals to be
approximated. It has been shown that residual error of the SMP
approximation decays exponentially for correlated signals with the
same support and additive white noise \cite{Tropp}.

\subsection{Spherical subspace selection with SMP}\label{sec:SSMP}

We propose to perform the classification of 3D face by
dimensionality reduction on the sphere. We therefore project the
3D point cloud onto the unit sphere $S^2$, and then we select a
subspace that spans functions on $S^2$. Since faces are typically
star-shaped objects, spherical projection preserves the face
geometry information, while reducing the classification complexity
by mapping a 3D signal to a 2D spherical signal. Each face, given
by a 3D point-cloud $\{p_n\}=\{(x_n,y_n,z_n)\}$ is, therefore,
represented as a spherical function $r=s(\theta,\varphi)$ sampled
at points  $\{(r_n,\theta_n,\varphi_n)\}$, which are obtained by
transforming Euclidean coordinates from the point cloud to
spherical coordinates given by ($\theta,\varphi$) that represent
the elevation and azimuth angles.

\begin{figure}[t]
\begin{center}
\includegraphics[width=0.45\textwidth]{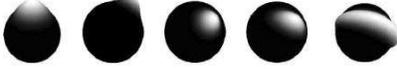}
\end{center}
\caption{Gaussian atoms.\label{fig:atoms}}
\end{figure}

Since we represent 3D faces as square-integrable functions on
$S^2$, denoted as $L^2(S^2)$, we can use the SMP to select a
subspace of spherical basis functions as a dimensionality
reduction step. We use a spherical dictionary proposed in
\cite{Tosic}, where the atoms are created by applying local
geometric transforms to a generation function $g(\theta, \varphi)$
defined on the sphere. Local transforms include atom motion
$(\tau, \nu)$ (position on the sphere with respect to $(\theta,
\varphi)$, respectively), rotation  $\psi$, and anisotropic
scaling by two scales $(\alpha,\beta)$ in orthogonal directions.
Motion and rotation are realized using a rotation in $SO(3)$,
which is the rotation group in $\mathbb{R}^3$. Five transform
parameters form the atom index $\gamma = (\tau, \nu, \psi, \alpha,
\beta ) \in \Gamma$, and the redundant dictionary is finally
constructed by applying a large set of different $\gamma$'s to
$g$. A detailed explanation of the dictionary construction is
given in \cite{Tosic}. An example of the generating function is a
2-D Gaussian function in $L^2(S^2)$, given by:
\begin{equation}
g(\theta, \varphi) = \exp (-\tan^2 \frac{\theta}{2}).
\label{eq:gen}
\end{equation}
Function in Eq.(\ref{eq:gen}) represents an isotropic gaussian
function, centered at the North Pole. In Figure \ref{fig:atoms} we
show a few sample Gaussian atoms that are obtained by applying
different local transforms to the generating function in
Eq.(\ref{eq:gen}).

Equipped with the spherical dictionary, we can directly apply SMP
to find the common support of the spherical faces, where the inner
product between two spherical functions $f=f(\theta,\varphi)$ and
$g=g(\theta,\varphi)$ is however given by:
\begin{equation}
\langle f, g \rangle = \int_{\theta} \int_{\varphi} f(\theta,\varphi) g(\theta,\varphi) \sin \theta d\theta d\varphi.
\label{eq:inner}
\end{equation}

In the following, we refer to this special case of SMP for
spherical signals using the dictionary defined on the sphere, as
\textit{simultaneous spherical matching pursuit} (SSMP).

\subsection{Recognition on the sphere}\label{sec:recog}

\begin{figure*}[t]
\begin{center}
  \includegraphics[width=0.8\textwidth]{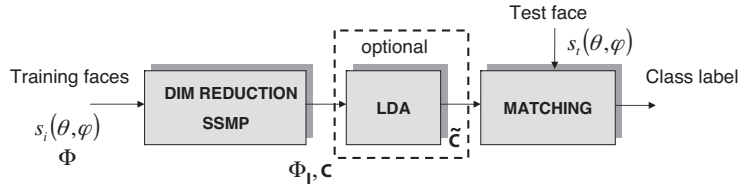}
  \caption{Block diagram of the recognition process.}
  \label{fig:recognitiondiagram}
\end{center}
\end{figure*}

The algorithm for recognition of 3D faces on the sphere is finally
illustrated in Figure \ref{fig:recognitiondiagram}. The first step
performs dimensionality reduction, by projecting the spherical
signals on the subspace spanned by the selected atoms i.e.,
span$\{ \mathcal{D}_I \}$, as described above. If we denote the
set of face signals by $S = [s_1,\ldots,s_n]$, the SSMP performs
the dimensionality reduction step by greedily selecting a set of
$K$ basis vectors $\mathcal{D}_I=
\{g_{\gamma_1},\ldots,g_{\gamma_K}\}$ from the dictionary
$\mathcal{D}$, such that all spherical faces are simultaneously
approximated as,
\begin{equation}
S \approx \Phi_I \cdot C.
\end{equation}
The matrix $C \in R^{K \times n}$ holds the
coefficient vectors (in its columns) and $\Phi_I= [g_{\gamma_1},
\ldots, g_{\gamma_K} ]$.

The coefficient vector conveys quite discriminative information
about the faces signals. However, the class separability of the
coefficient vectors in the reduced space could yet be improved by
performing an optional Linear Discriminant Analysis (LDA) step
before matching. LDA exploits the class labels information of the
training samples in order to enhance the discriminant properties
of the coefficient vectors. It introduces supervision in the
recognition process and permits to build a new set of coefficient
vectors $\tilde{C} = C W$ where the weights $W$ are chosen to
optimize the ratio of between-class variance and within-class
variance for training data \cite{Belhumeur}.

Finally, the matching is performed by comparing the coefficient
vectors $C$, which represent the lower dimensional data samples.
The recognition is performed by nearest neighbor classification.
We iteratively compute the coefficients $c_t$ of the test face
signal $s_t$ on the sub-dictionary $\mathcal{D}_I$. The
classification is then performed by computing the $L_1$ distance
between $c_t$ and any coefficient vector $c_i$ corresponding to
the training signals
\begin{equation}
d(c_t, c_i) = \sum_{j = 1}^{K}|c_t(j)-c_i(j)| .
\label{eq:distL1}
\end{equation}
The class of the test signal is finally given by the class of the
signal $s_i$ that leads to the smallest distance $d(c_t, c_i)$
between the coefficients vectors. The same classification method
is used for coefficients $\tilde{C}$ modified by LDA. The choice
of the $L_1$ distance metric is mostly empiric as it leads to
superior classification performance compared to other metrics.


\section{Experimental results}\label{sec:exp}

\subsection{Experimental setup}

In this section, we evaluate the performance of the proposed
algorithms in both recognition and verification scenarios. We
compare our algorithms with PCA and LDA on depth images that  have
undergone the same preprocessing step as the data used in the SSMP
algorithm. PCA and LDA are well known methods that represent
state-of-the-art technologies for 3D recognition.

For our evaluation, we use the UND (University of Notre Dame)
Biometric database \cite{UND1,UND2}, also known as
FRGC v.1.0 database. It contains 953 facial images of 277 subjects, where each subject
has between one and eight scans. Each facial scan is provided in
the form of a 3D point-cloud, along with a corresponding binary
matrix of valid points. The number of vertices in a point-cloud
typically varies between 30.000 and 40.000.

\begin{table}
\begin{center}
\begin{tabular}[t]{|c|c|c|c|c|}
  \hline
  Test & $i$ & Number of  & Training & Test \\
  configuration & & subjects & set & set \\   \hline
  $T_1$ & 1 & 200 & 200 & 673\\   \hline
  $T_2$ & 2 & 166 & 332 & 474\\   \hline
  $T_3$ & 3 & 121 & 363 & 308\\   \hline
  $T_4$ & 4 & 86 & 344 & 187\\   \hline
\end{tabular}
\end{center}
\caption{Test configurations and their characteristics.}
\label{tab:configs}
\end{table}

We defined several test configurations for our experimental
evaluation. Each configuration is characterized by the number of
samples per subject that form the training set. For each
configuration $T_i$, we keep only the subjects from the database
that have at least $i+1$ samples, and we use $i$ training samples
per class (randomly chosen), while assigning the rest to the test
set. The subjects that have only one facial scan can not
be used in the recognition tests. Table \ref{tab:configs}
summarizes the test configurations and their main characteristics.

\paragraph{SSMP implementation}
For the dictionary construction in SSMP-based methods, we have
used the 2D Gaussian on the sphere (\ref{eq:gen}) as the
generating function. The atom indexes $\gamma$ that define the
dictionary, have to take discrete values in practice.  We use here
a discretization of the dictionary as in~\cite{Tosic}, mostly
built on empirical choices for atom parameter values. The position
parameters, $\tau$ and $\nu$ are uniformly distributed on the
interval $[0, \pi]$, and $[-\pi,\pi)$, respectively, with equal
resolution of 128 points. The rotation parameter $\psi$ is uniformly
sampled on the interval $[-\pi, \pi)$, with the same resolution as
$\tau$ and $\nu$. This choice is mostly due to the use of fast
computation of correlation on SO(3) for the full atom search
within the SSMP algorithm. In particular, we used the
\emph{SpharmonicKit}
library\footnote{\url{http://www.cs.dartmouth.edu/~geelong/sphere/}},
which is part of the \emph{YAW
toolbox}\footnote{\url{http://fyma.fyma.ucl.ac.be/projects/yawtb/}}.
Finally, scaling parameters are distributed in a logarithmic
manner, from $1$ to half of the resolution of $\tau$ and $\nu$,
with a granularity of one third of octave. The largest atom covers
half of the sphere.

The use of fast computation of correlation on the SO(3) group
requires the spherical data  to be sampled on an equiangular
($\theta,\varphi$) grid, defined as:
\begin{equation}
G=\{(\theta_i,\varphi_j),
\theta_i=\frac{(2i+1)\pi}{2N_\theta},\text{ and }
\varphi_j=\frac{j2\pi}{N_\varphi}\}.
\end{equation}
where: $i=0,...,N_\theta-1$ and  $j=0,...N_\varphi-1$. Since 3D
face point clouds are projected as scattered data on the sphere,
an interpolation step is necessary. For its simplicity we use
k-nearest neighbor interpolation, where the value on each
spherical grid point $(\theta_i,\varphi_j)$ is computed as an
average of its $k$ nearest neighbors. We have used $k=4$ and a
resolution of $N_\theta=128, N_\varphi=128$. Note finally that,
for the sake of computational ease, dimensionality reduction with
SSMP is performed off-line, using only one training face per
subject. The resulting subspace is then used for projecting both
training and test samples.

\paragraph{Virtual faces}
The size of the training set is important in determining the
classification performance. We propose to enrich the training set
with \emph{virtual faces} (see e.g., \cite{decoste2000dir} and
references therein). These are faces that are artificially
generated by slight variations of the original training faces.
They are given the corresponding class labels of the training face
they originate from, and they are treated as training samples. The
use of virtual faces is motivated by two main reasons: (i) they
compensate for small registration errors (recall that our
registration process is fully automatic and it is expected to
contain a few registration errors) and (ii) by augmenting the
training set, they may contribute to the performance of
sample-based methods (e.g., LDA) that can benefit from large
sample sets. Note that the virtual faces do not introduce any new
information to the training set, since they are synthetically
generated by the original training faces. For computational
convenience, we construct them by one or two pixel translations in
the spherical domain. Note finally that virtual faces are used
only in the SSMP+LDA method.

\subsection{Recognition results}

We present recognition results of our methods and we compare them
with PCA and LDA on depth images. For the sake of completeness, we
also report the classification performances of the Euclidean
distance (EUC) between depth images, and Mean Square Error (MSE)
between spherical functions. For the two latter methods, each test
face is recognized as the closest neighbor in the training set. In
SMMP+LDA (resp. PCA+LDA), the number of dimensions used in LDA is
set to the minimum between the number of features in SSMP (resp.
PCA) and $c-1$, where $c$ is the number of classes (subjects). Virtual faces are
used in the SSMP+LDA method in configurations $T_1$, $T_2$ and
$T_3$ only, since they correspond to small training sets. In these
cases, each training face is used to generate 8 virtual
faces.

We start with rank-1 recognition, which refers to the scenario
where a class prediction is considered to be a hit when the label
of the closest neighbor is the correct one. Then, we will discuss
the generic rank-$k$ scenario, where the prediction is a hit when
the correct label is included in the labels of the closest $k$
neighbors.

\paragraph{Rank-1 recognition}\label{sec:rank1recognition}

\begin{figure*}[tb]
\begin{center}
\subfigure[Test Configuration $T_1$]{
    \includegraphics[width=0.45\textwidth]{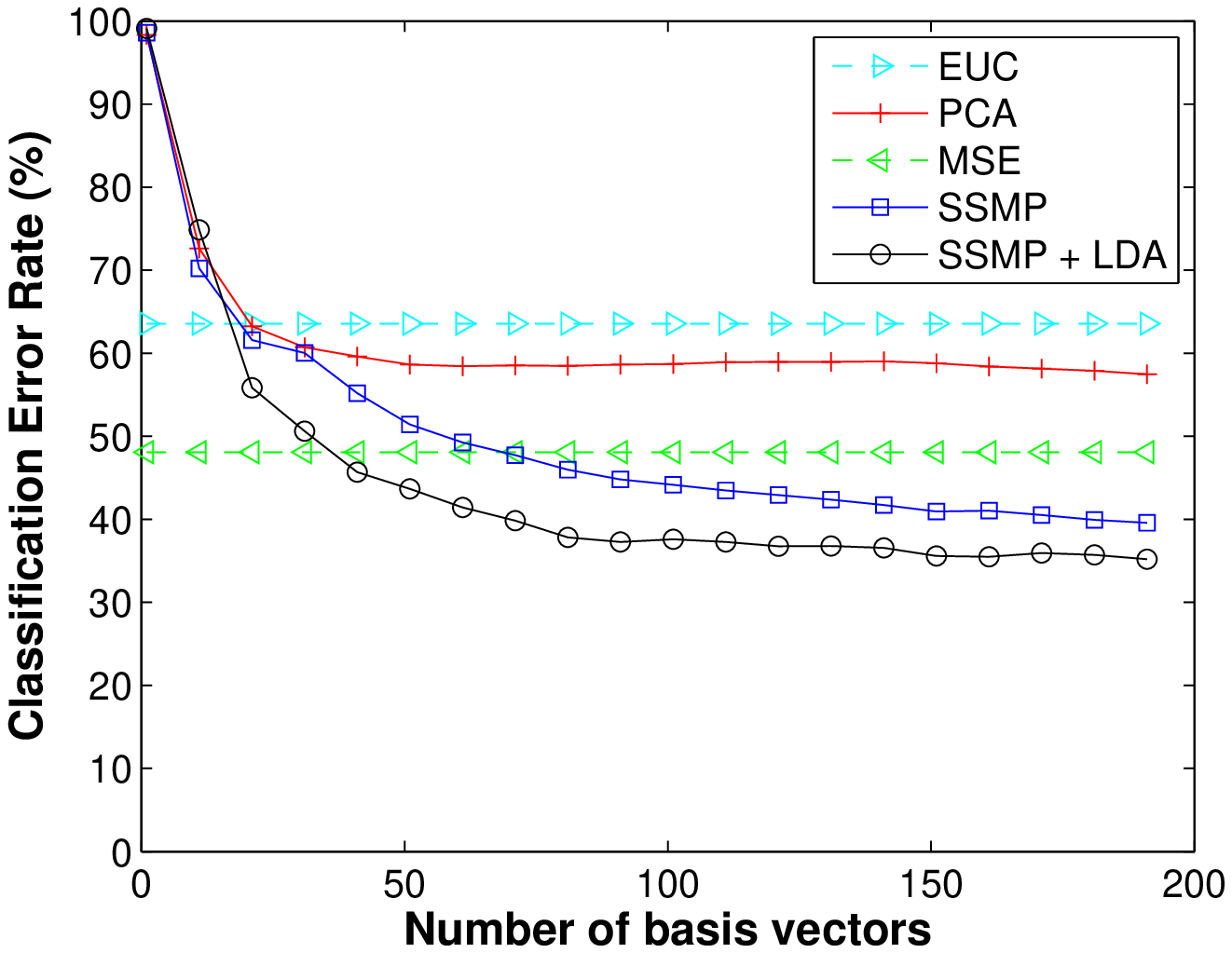}}
\subfigure[Test Configuration $T_2$]{
    \includegraphics[width=0.45\textwidth]{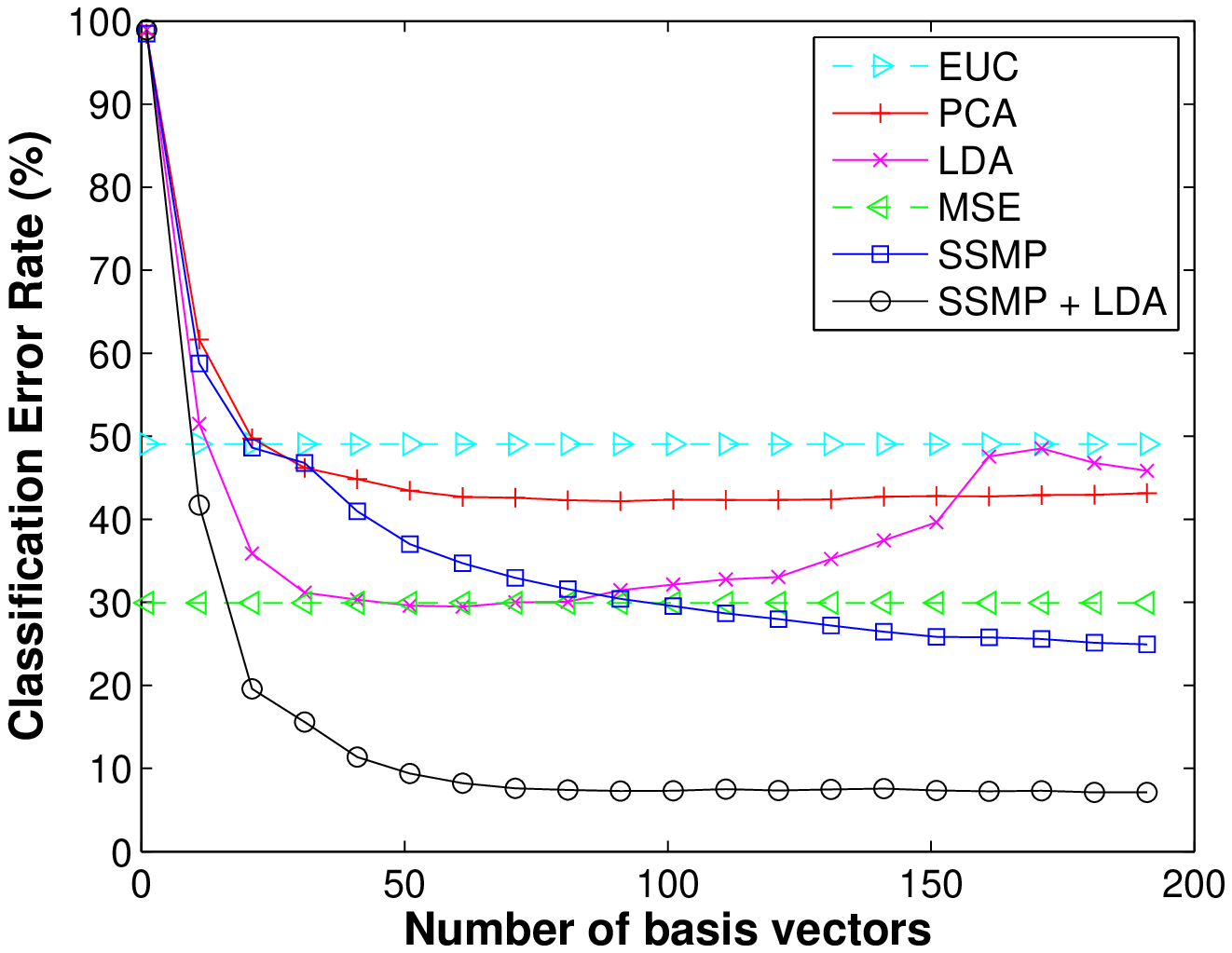}}\\
\subfigure[Test Configuration $T_3$]{
    \includegraphics[width=0.45\textwidth]{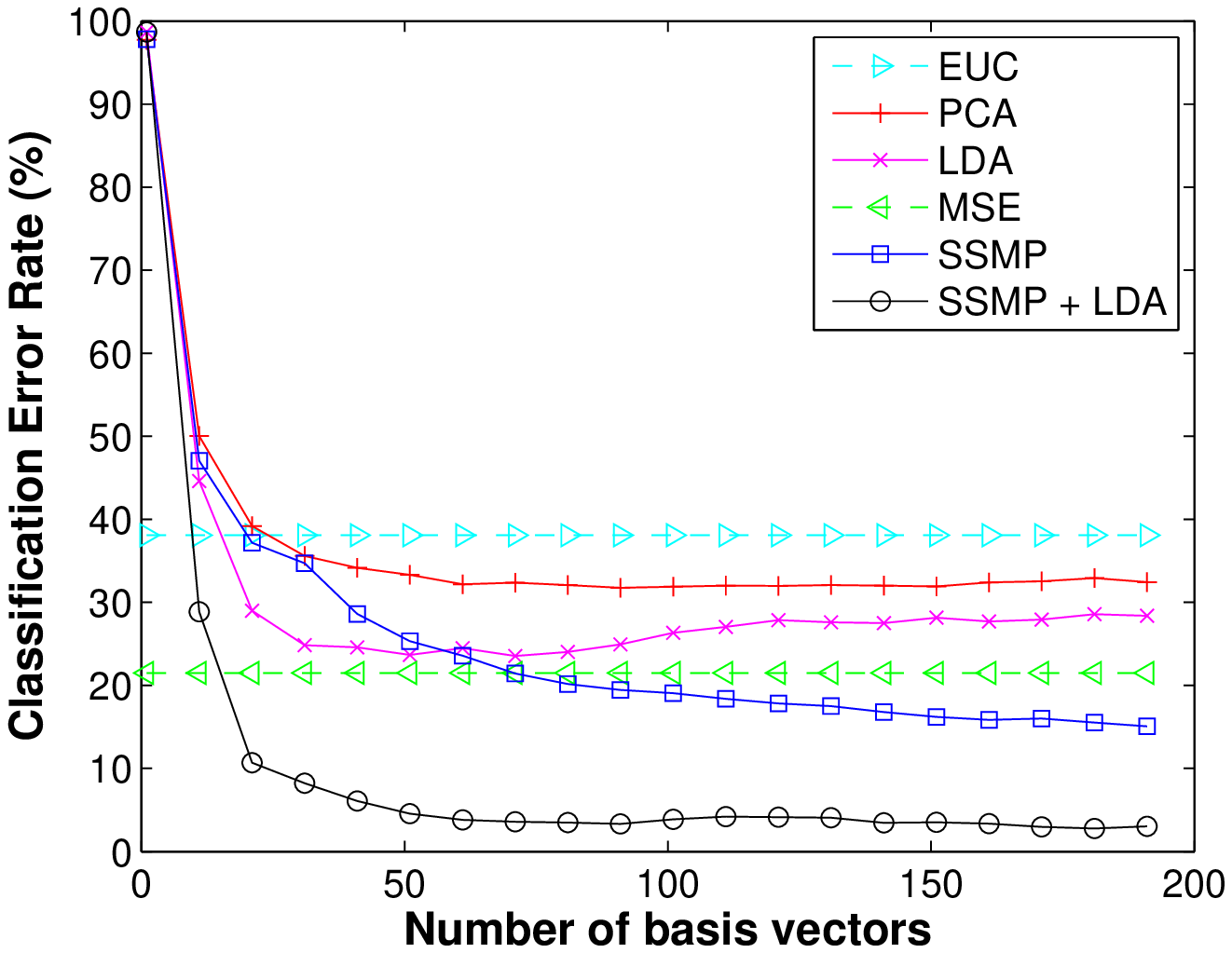}}
\subfigure[Test Configuration $T_4$]{
    \includegraphics[width=0.45\textwidth]{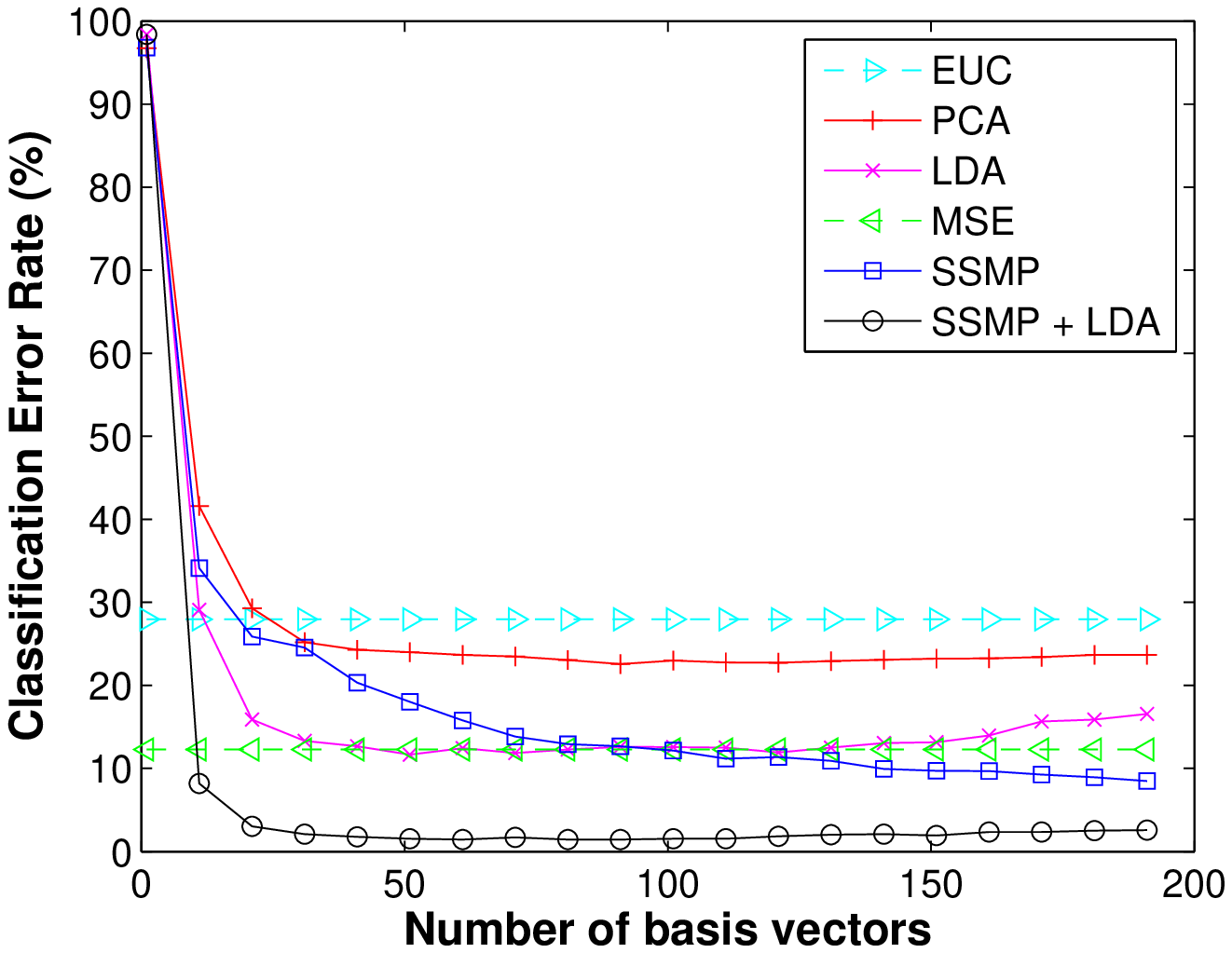}}\\
\end{center}
\caption{Rank-1 recognition results: average classification error
rate versus the dimension of the
subspace.\label{fig:recognitionRank1} }
\end{figure*}

\begin{table}[tb]
\begin{center}
\begin{tabular}[b]{|c|c|c|c|c|c|c|}
  \hline
                   & $T_1$ & $T_2$ & $T_3$ & $T_4$      \\   \hline
  PCA              & 45,17 & 60,97 & 74,35 & 82,89      \\  \hline
  PCA + LDA        & -     & 74,89 & 80,52 & 93,58         \\   \hline
  SSMP             & 62,85 & 77,22 & 87,01 & 94,12      \\   \hline
  SSMP + LDA       & 67,61 & 94,73 & 98,70 & 100            \\   \hline
\end{tabular}
\caption{Best rank-1 recognition rates (\%) reached by each method
in experiment \ref{sec:rank1recognition}. \label{tab:results}}
\end{center}
\end{table}


All tests are performed 10 times, by splitting randomly the
samples into the training and the test sets. Figure
\ref{fig:recognitionRank1} shows the classification error rate for
all configurations, averaged over the 10 random experiments.
Notice the remarkable improvement introduced by the employment of
spherical functions for facial representation. This is evident
from the fact that the recognition performance of nearest neighbor
classification with Mean Square Error (MSE) between spherical
signals, outperforms that of Euclidean distances between depth
images (EUC). This provides also the main motivation for working
on the sphere. Based on this observation, it seems reasonable that
our SSMP algorithm outperforms PCA in all configurations. Notice
finally that SSMP+LDA is the best performer. In T2, SSMP reaches
recognition performance of $77,22\%$, while SSMP+LDA reaches
$94,73\%$. The latter goes to the maximum $100\%$ in T4, even in the absence of virtual faces. Table
\ref{tab:results} shows the highest recognition rates achieved by
each method in all configurations.

\paragraph{Rank-$k$ recognition}
\begin{figure*}[tb]
\begin{center}
\subfigure[Test Configuration $T_1$]{
    \includegraphics[width=0.45\textwidth]{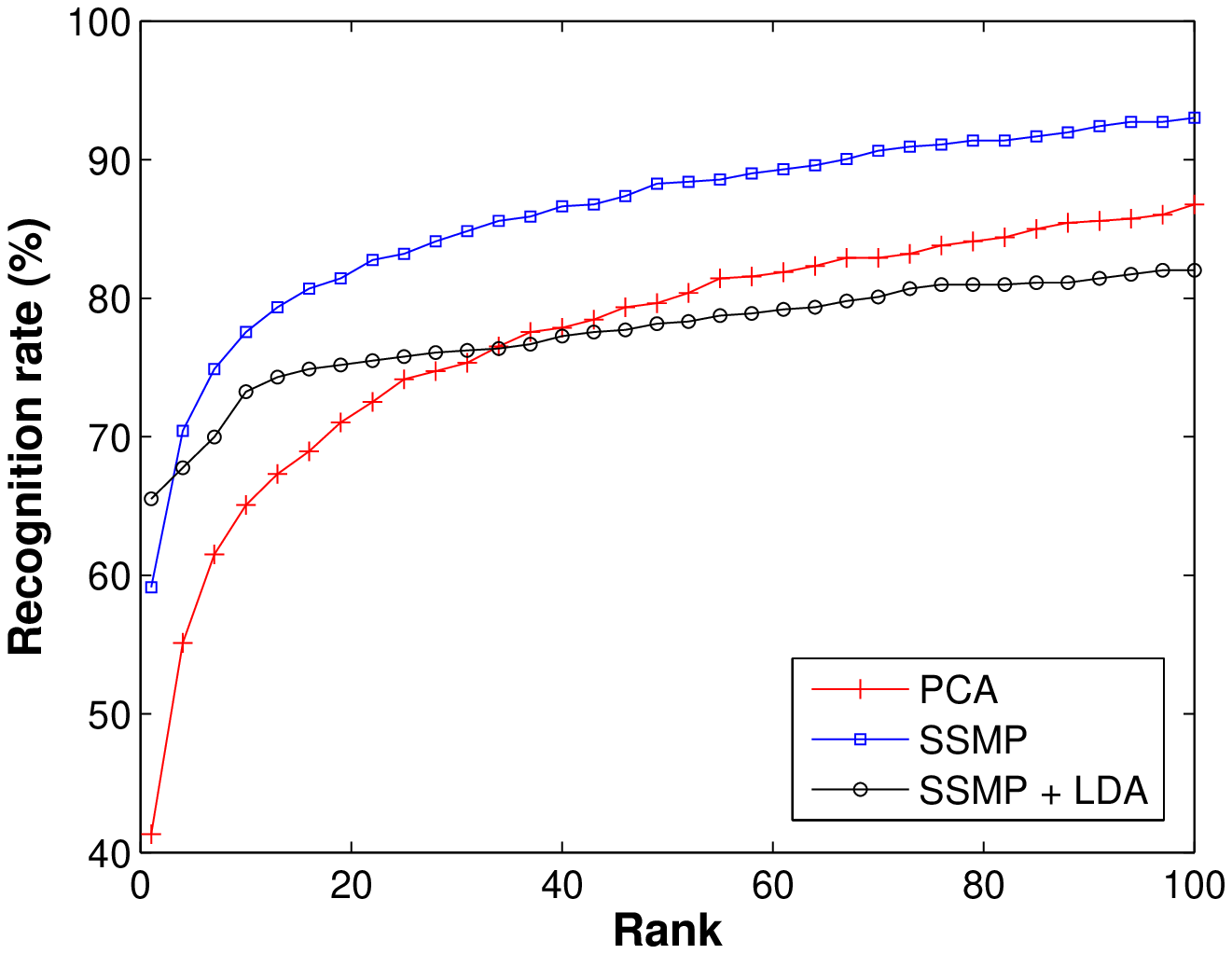}}
\subfigure[Test Configuration $T_2$]{
    \includegraphics[width=0.45\textwidth]{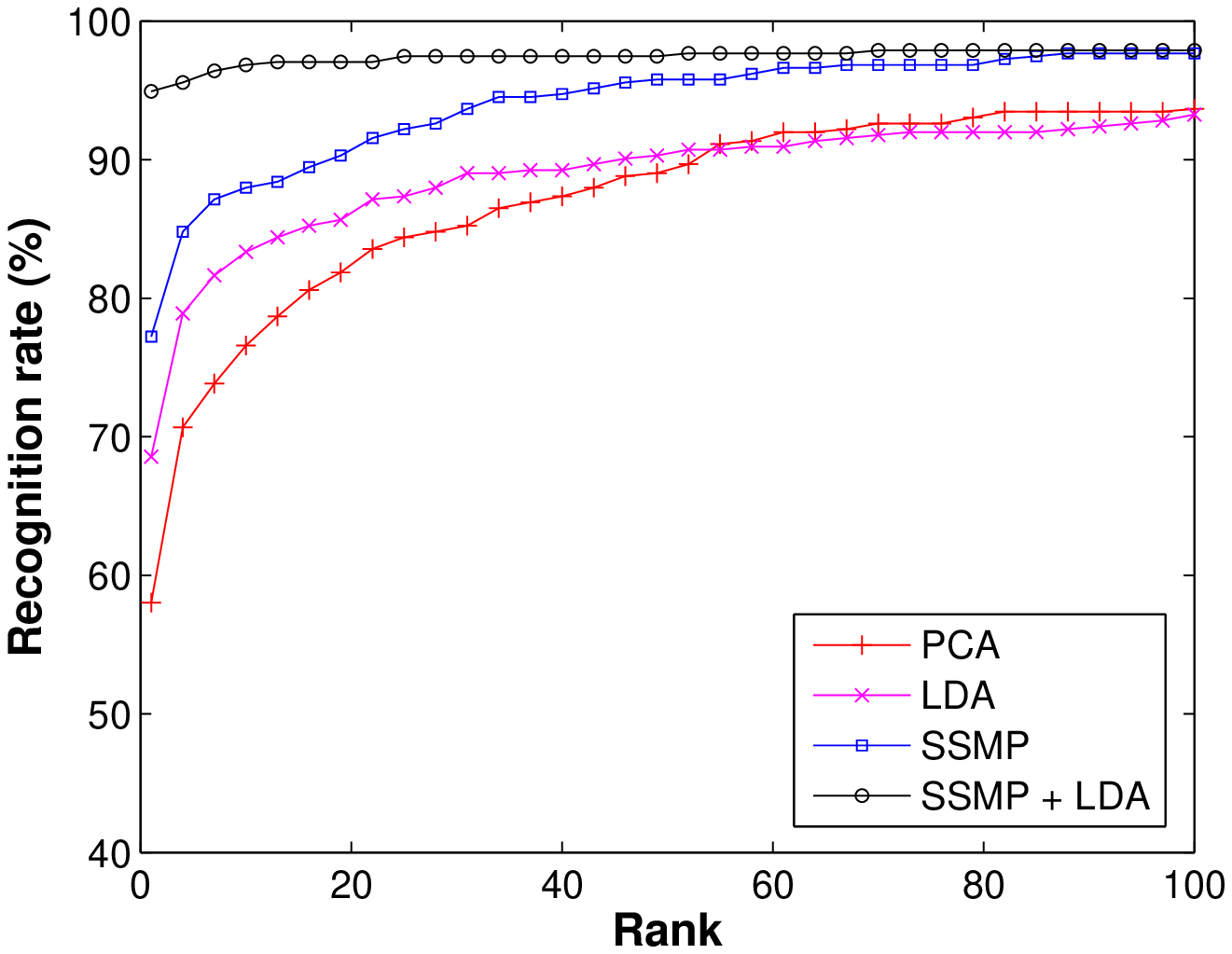}}\\
\end{center}
\caption{Rank-$k$ recognition results in terms of CMC
curves.\label{fig:recognitionRankK} }
\end{figure*}

We report rank-$k$ recognition performances in terms of cumulative
match characteristic (CMC) curves. A CMC curve simply illustrates
the fluctuation of the recognition rate versus the rank $k$.
Figure \ref{fig:recognitionRankK} shows the obtained CMC curves
for $T_1$ and $T_2$ that represent the most interesting cases,
since T3 and T4 correspond to very good performances for all
methods. The CMC curves in this figure are averages over 10 random
tests, where the best number of dimensions for each algorithm is
used (obtained from the previous rank-1 recognition experiments).
As expected, notice again that SSMP is superior to PCA, and LDA
introduces in both methods a significant performance boost.

\subsection{Verification results}

\begin{figure*}[htb]
\begin{center}
\subfigure[Test Configuration $T_1$]{
    \includegraphics[width=0.45\textwidth]{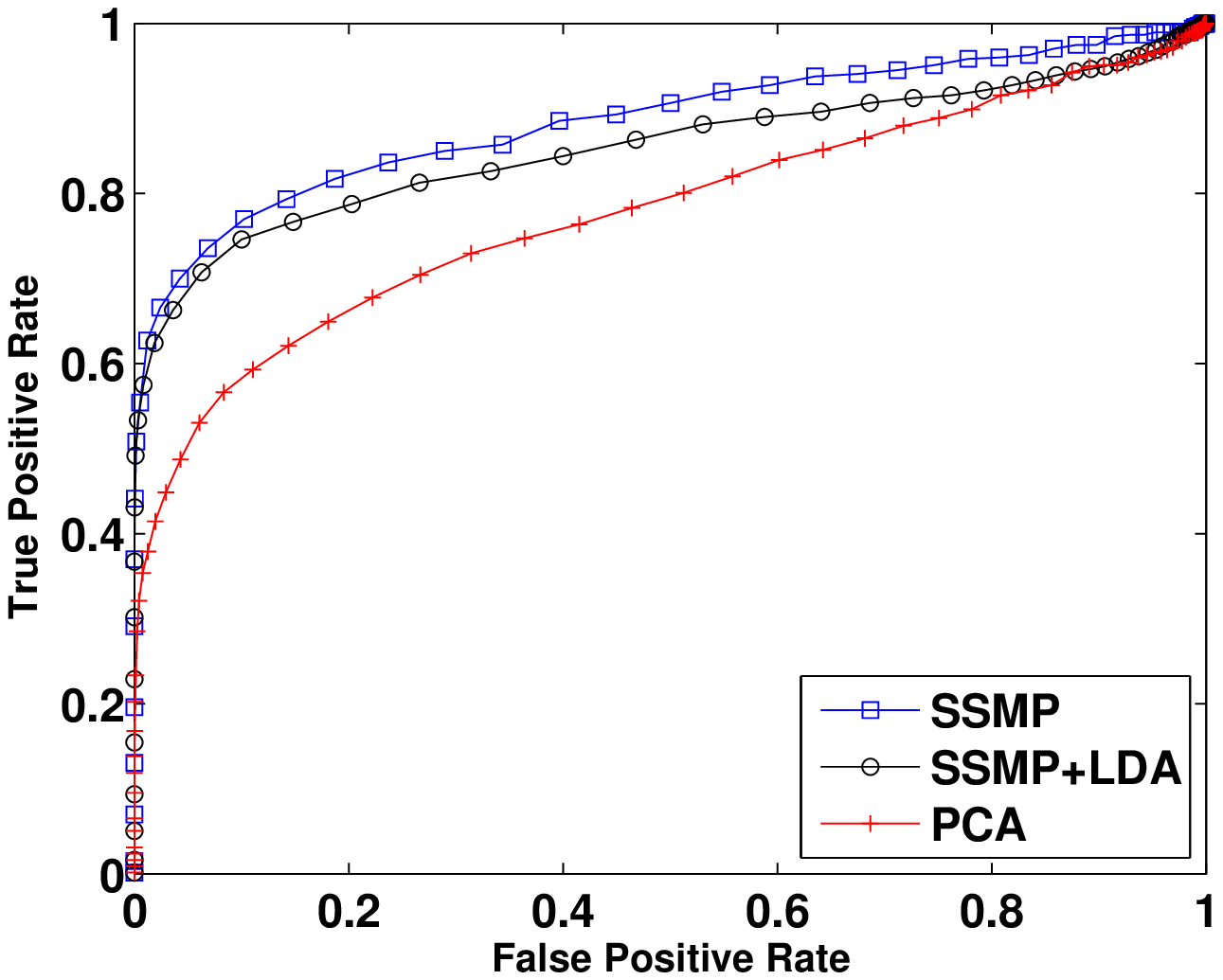}}
\subfigure[Test Configuration $T_2$]{
    \includegraphics[width=0.45\textwidth]{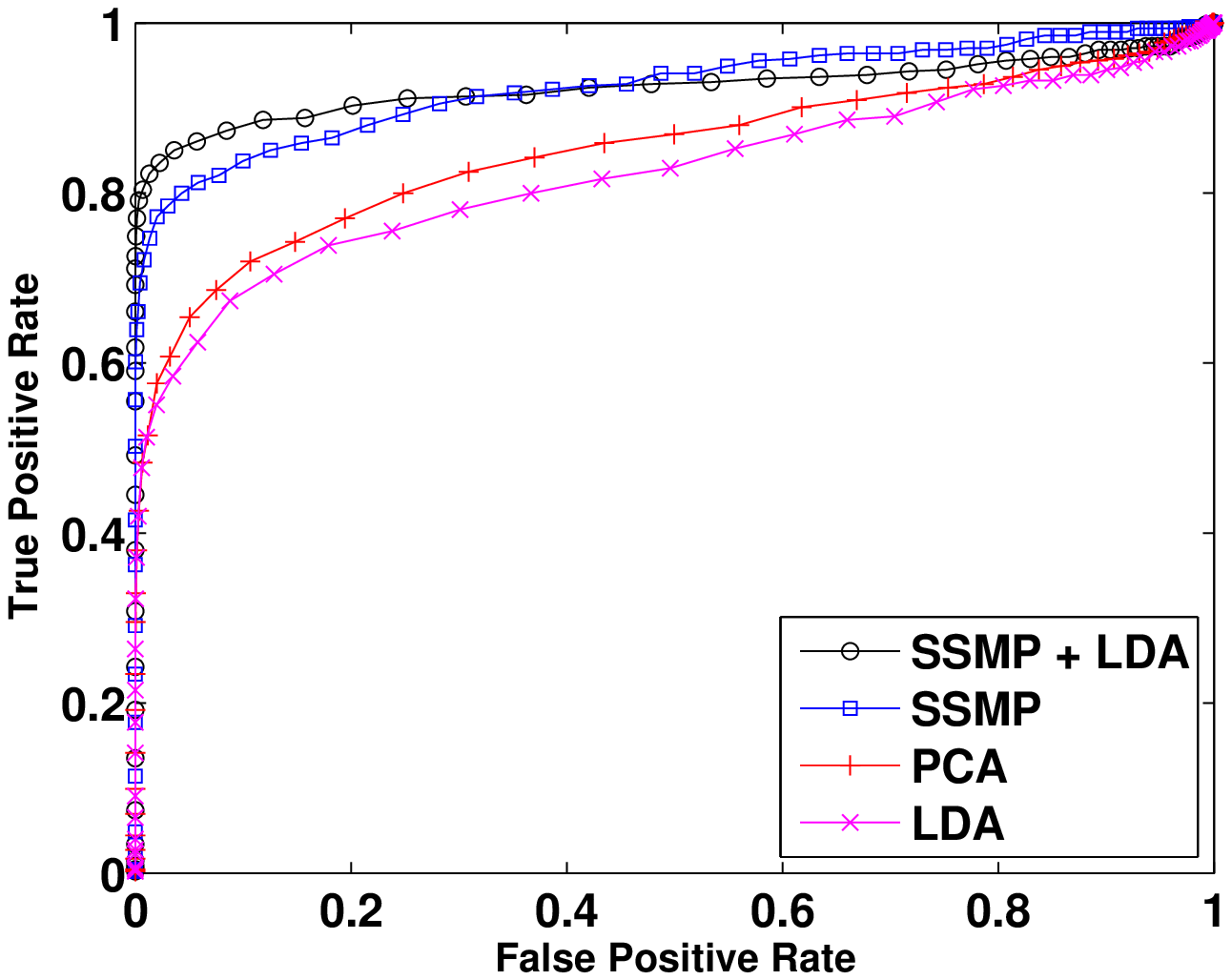}}\\
\subfigure[Test Configuration $T_3$]{
    \includegraphics[width=0.45\textwidth]{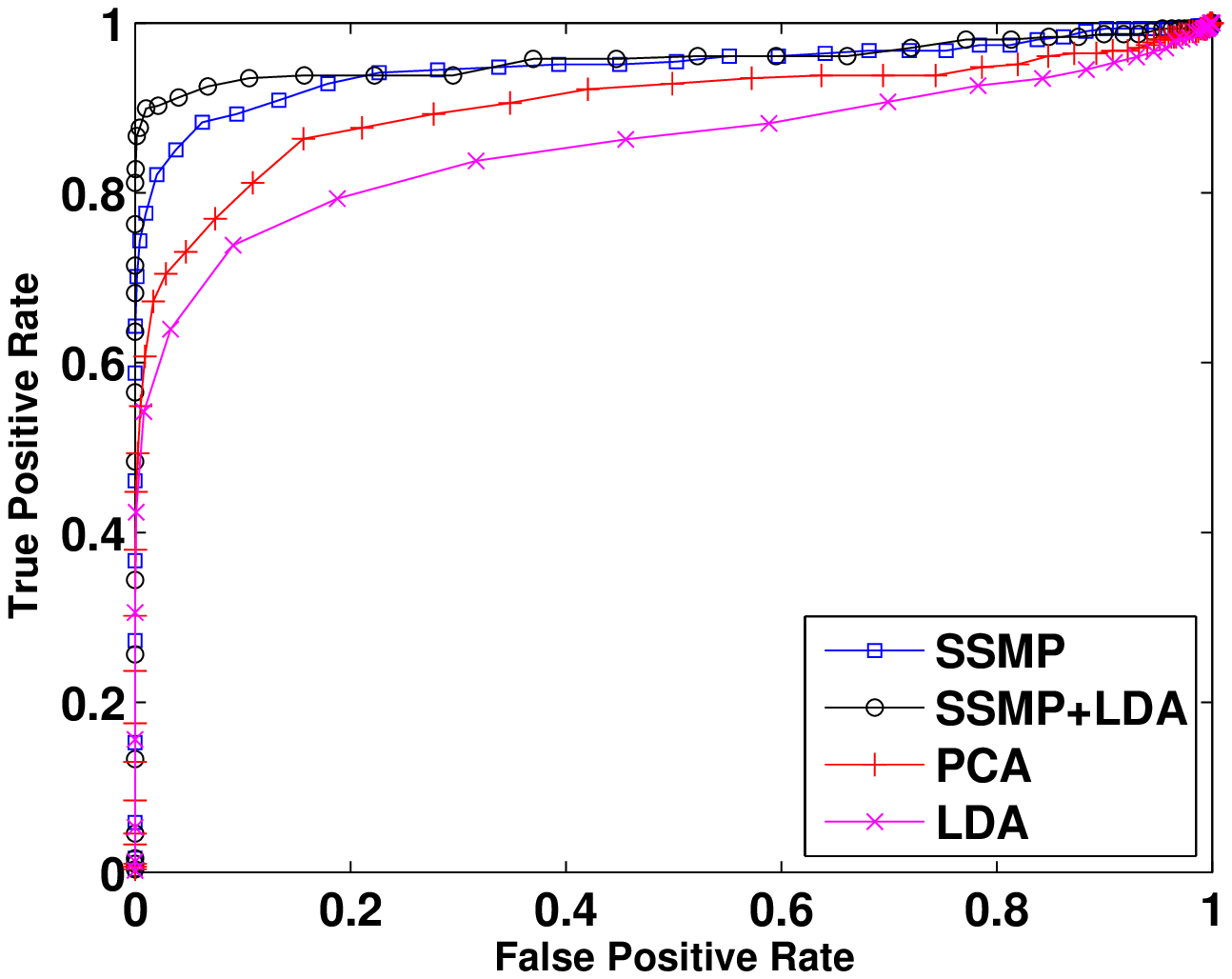}}
\subfigure[Test Configuration $T_4$]{
    \includegraphics[width=0.45\textwidth]{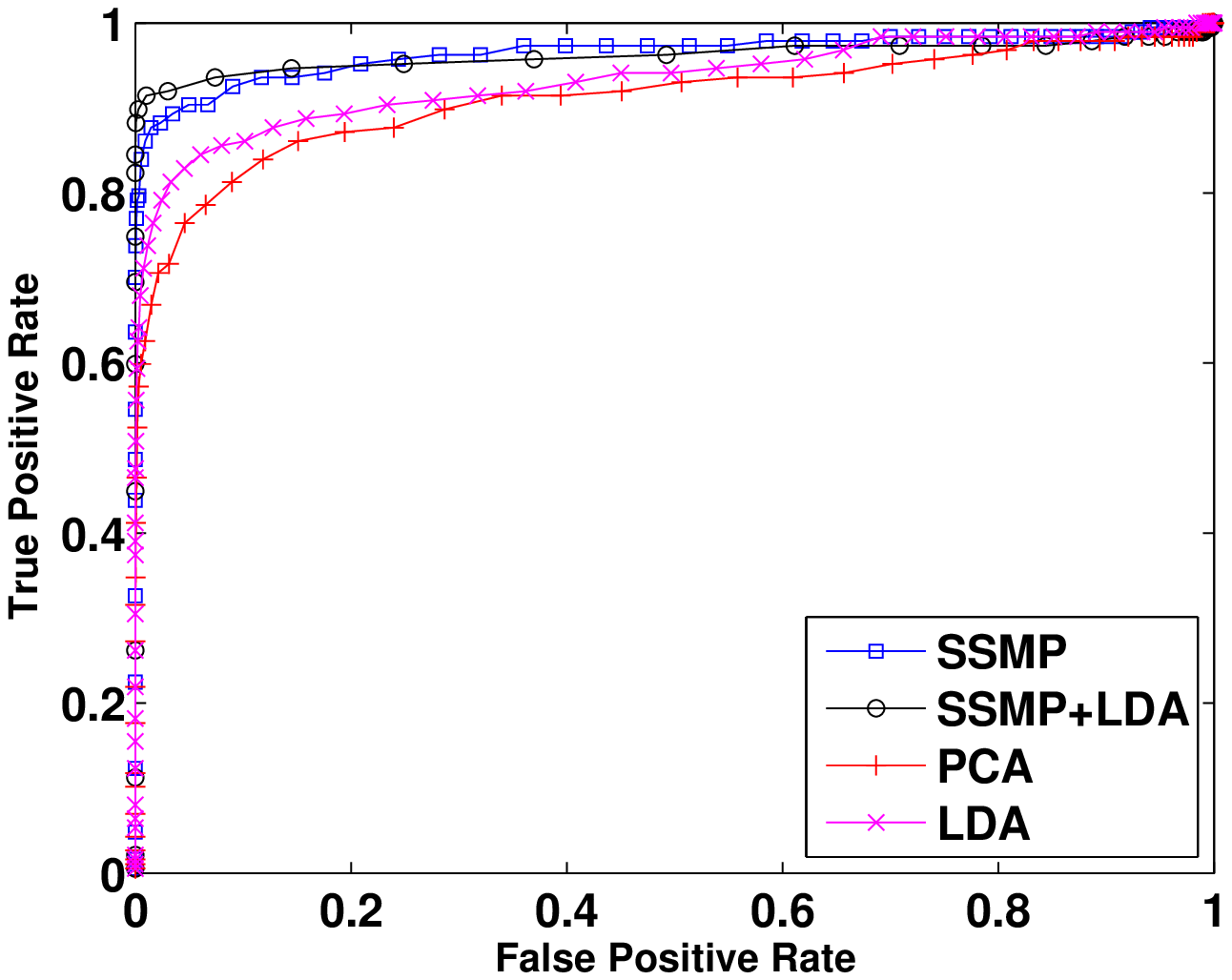}}\\
\end{center}
\caption{Verification performance in terms of ROC
curves.\label{fig:verification}}
\end{figure*}

We compare now all the above methods in the verification scenario,
where the test subject claims an identity and the system has to
either accept or reject this claim. If the identity is the correct
one, then the test subject is called a \textit{client}; otherwise,
it is called an \textit{impostor}. In systems that output a
confidence score about the test subject, a hard decision (i.e.,
accept or reject) is typically reached according to a threshold
value. We report the verification performances in terms of
receiver operating characteristic (ROC) curves, which show the
fluctuation of the true positive rate (TPR) versus the false
positive rate (FPR) across all values of the threshold. For the
computation of the ROC curve we consider every possible pair of
subject and claimed identity.

In our experimental setup, we use the dimensions that yields the
best performance, which corresponds to 200 atoms in SSMP and 100
dimensions in PCA. The number of LDA dimensions in both SSMP+LDA
and PCA+LDA is set with the same rule as in the recognition
experiments (i.e., using the minimum between the number of
PCA/SSMP features and $c-1$). Also, in SSMP+LDA we use virtual
faces only for configurations T1 and T2. Figure
\ref{fig:verification} shows the average ROC curves over 10 random
experiments for all configurations. Similar conclusions can be
drawn here as well. Unsurprisingly, observe again that SSMP
consistently outperforms PCA in all configurations and SSMP+LDA is
the best performer.

\subsection{Discussion}

It is worth noting that supervised versions of SSMP could be also
used \cite{Efi}. The idea would be then to select the atoms from
the dictionary according to discriminative criteria. However, in
the proposed scheme the supervision information is already taken
into account in the LDA postprocessing step, and prior experience
has shown that this suffices, when predefined dictionaries are
used.

Note also that the importance of each region of the face in terms
of recognition performance is certainly not uniform \cite{Wong}.
Although the selection of such regions is typically performed
manually and it maybe sensitive to the testing conditions, one
possible approach to take advantage of this observation could be
to group the features selected by SSMP into regions by clustering
on the sphere, do a classification per region and then fuse the
results (e.g., by majority voting). Such an approach however
requires a sufficient number of atoms in each area, and the
performance of such a region-based classifier has not been
convincing.

Note finally that the proposed dimensionality reduction scheme is
generic and simple extensions could be proposed to make the
classification more sensitive to some specific areas. For example,
the SSMP scheme can easily be adapted to give priorities to
regions of high interest such as the nose or the eyes. Such a
prioritization can be achieved by giving proper weights to atoms
located in different areas, in order to force the dimensionality
reduction step to select features in areas that are expected to be
more discriminative. This however goes along the lines of
supervised versions of SSMP mentioned above with the main
difference that discriminative capability in this case is mostly
defined in a region-based way.


\section{Conclusions}
We have proposed a methodology for 3D face recognition based on
spherical sparse representations. First, we introduced a fully
automatic process for extraction, preprocessing and registration
of facial information in 3D point clouds. Next, we proposed to
convert faces from point clouds to spherical signals. Sparse
spherical representation of faces allows for effective
dimensionality reduction through simultaneous sparse
approximations. The dimensionality reduction step preserves the
geometry information, which in turn leads to high performance
matching in the reduced space. We provide ample experimental
evidence that indicates the advantages of the proposed approach
over state-of-the-art methods working on depth images.

\section{Acknowledgements}
The authors would like to thank Prof. Patrick Flynn for sharing
with us the UND Biometrics database.

\end{document}